\documentclass[10pt,twocolumn,letterpaper]{article}

\usepackage{cvpr}
\usepackage{times}
\usepackage{epsfig}
\usepackage{graphicx}
\usepackage{amsmath}
\usepackage{amssymb}

\usepackage[pagebackref=true,breaklinks=true,letterpaper=true,colorlinks,bookmarks=false]{hyperref}

\cvprfinalcopy 

\def\cvprPaperID{****} 

\ifcvprfinal\fi
\begin{document}

\title{Learning to count with deep object features.}

\author{Santi~Segu\'{i},
        Oriol~Pujol,
        and~Jordi~Vitri\`{a}
\thanks{S. Segu\'{i}, O. Pujol and J Vitri\`{a} are with the Dept. Matematica Aplicada i Analisis, Universitat de Barcelona, Barcelona, Spain, and Computer Vision Center (CVC), Barcelona, Spain.}
}


\maketitle

\begin{abstract}
Learning to count is a learning strategy that has been recently proposed in the literature for dealing with problems where estimating the number of object instances in a scene is the final objective. In this framework, the task of learning to detect and localize individual object instances is seen as a harder task that can be evaded by 
casting the problem as that of computing a regression value from hand-crafted image features. In this paper we explore the features that are learned when training a counting convolutional neural network in order to understand their underlying representation. To this end we define a counting problem for MNIST data and show that the internal representation of the network is able to classify digits in spite of the fact that no direct supervision was provided for them during training. We also present preliminary results about a deep network that is able to count the number of pedestrians in a scene. 
\let\thefootnote\relax\footnote{This paper has been accepted at Deep Vision Workshop at CVPR 2015.}
\end{abstract}

\cvprsection{Introduction}
Counting the number of instances of an object in an image can be approached from two different perspectives: 1) training an object detector, and 2) training an object counter. In the first case we must provide the system with a large set of object examples, properly labeled and localized, that represent most of the possible views and appearances of the object, and the result is an object classifier. In the second case we only need to provide the number of object instances for each image sample and the result is typically a regressor \cite{Lempitsky2010}.

Until the success of convolutional neural networks (CNN), most object detectors where based on sophisticated classifiers that relied on hand-crafted features \cite{Felzenszwalb2010}.  These classifiers needed a relatively small number of samples to be trained. Labeling was provided by drawing rectangular bounding boxes on individual object instances. The success of this approach in real counting problems, such as the problem of counting people in crowded environments, was modest and for this reason several authors proposed the training of regression-based systems \cite{Lempitsky2010, Flaccavento2011}.

A counting system is a mapping from the image space to the set of non negative integers. There are two choices when defining this kind of systems: the image features and the mapping function.  

Image features can be defined at object level or can be more abstract features that are good surrogates of the number of object instances. For example, in the case of people counting in static environments, these features can represent parts of the background that can be occluded by people or features that are related to the appearance of people groups.

If the number of instances is bounded and we have available samples for all cardinalities, the mapping function can be implemented by a multiclass classifier that directly predicts the number of object instances in the scene. Otherwise, we must rely on multivariate regressors that learn a mapping from the image feature space to the (bounded) set of positive integers. Classical regression functions are prone to errors when the input is high dimensional and for this reason more advanced methods such as Gaussian Process regression and Bayesian regression have been recently proposed \cite{Chan2012}.

The remarkable performance of CNN at a number of different vision tasks suggests the use of this technique to learn to count objects in a scene. Several advantages can be foreseen from this application, being the most important that of learning image features from samples instead of hand-crafting highly specialized image features that are dependent on the object class. 

But there could be some potential advantages that go beyond the feature issue. For example, CNN have shown their capacity of knowledge transfer for a number of tasks or the ability of simultaneously performing different tasks even when trained for only one \cite{Zhou2014}. What kind of tasks can be indirectly learnt by counting CNN? Can "learning to count" be a surrogate for fully supervised learning?

\begin{figure*}[!ht]
\centering
\includegraphics[height=7cm]{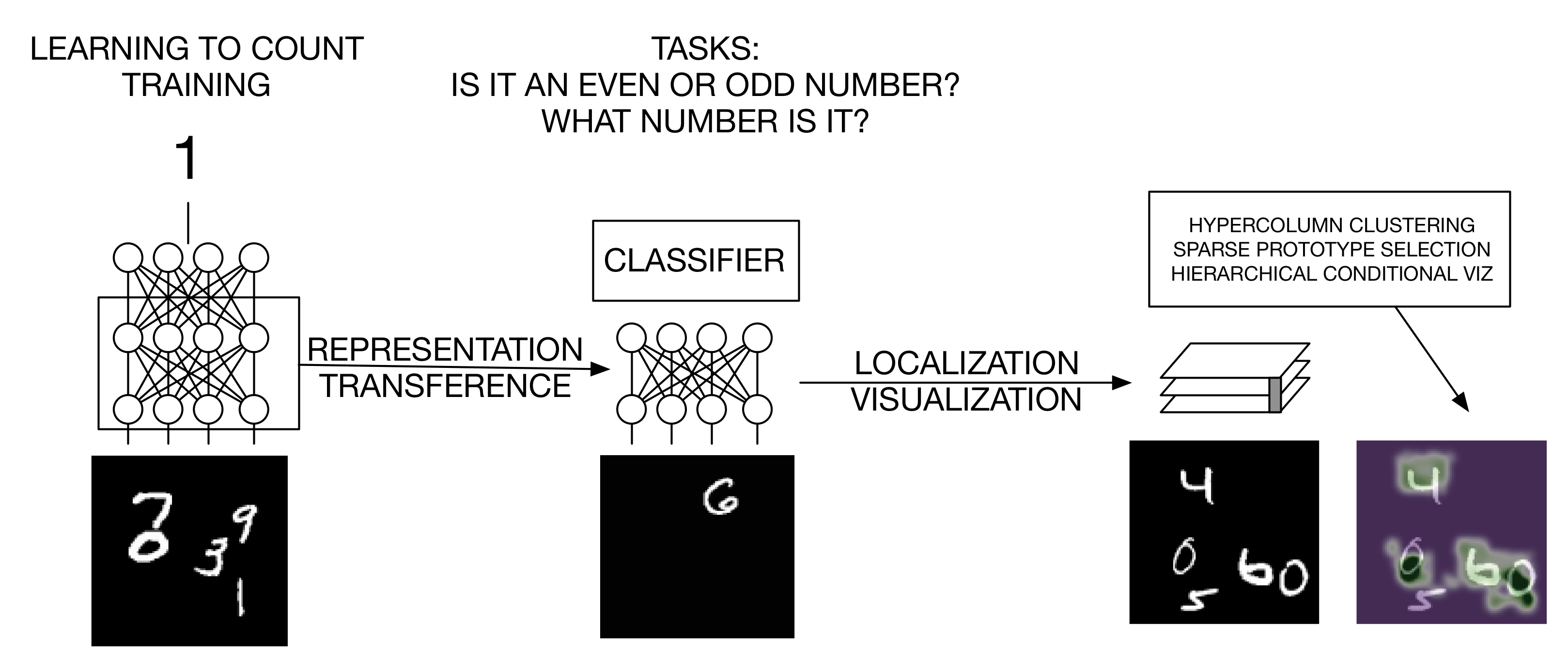}
\caption{Learning to count can be seen as a weakly supervised learning problem that when trained with CNN produces good object representations. In this paper we show that the features of a CNN that has been trained to count digits can be readily used for more specific classification problems and even to localize digits in an image.}
\label{fig:learning_to_count_concept}
\end{figure*}

\cvprsubsection{Learning to count from the point of view of supervision}

Fully supervised learning requires exhaustive labelling of the concept of interest. However, when the labeling cost is too high this becomes prohibitive. In order to soften this cost, several strategies have been proposed in literature. For example, semi-supervised learning \cite{Chapelle2006} takes advantage of few fully labeled data and large amounts to unlabeled data. Another way of decreasing the annotation task effort is by means of weakly supervised labelling. Under this name, one finds different definitions of the term "weakly". In some works, it is a surrogate for the concept of noisy labels \cite{Dekel2009}, i.e. labels provided by different supervisors with different quality. In computer vision, we also find the concept of "weakly" labeled data associated to two different tasks, either imperfect annotations\cite{Raykar2009} or for only indicating the presence of an object in an image\cite{Wang2013}. In the last case, the only information given to the learning process is the presence or absence of objects but not their location.



Following this line of work, in this paper we explore "learning to count" as a further generalization of weakly supervised learning. Different to multiple instance learning\cite{Foulds2010}, positive and negative bags are not provided but a generalization in terms of the number of occurrences of the object of interest. In this sense, we hypothesize that the problem of learning of the concept of interest could be casted in some cases as a problem of "learning to count" without explicitly giving information about what we are counting. Besides the straightforward application of this process to multiple occurrence counting, our hypothesis is that by using this new weak indirect annotation, we may be able to infer discriminant features for the identification of the concept of interest.

From the point of view of annotation our proposal lies in the middle of weakly supervised and fully supervised approaches. Whereas in fully supervised learning the object boundary or bounding box is given to the learning process in this proposal we only require the multiplicity of the concept of interest to be given. This saves annotation effort. Similar to weakly supervised techniques the location of the concept of interest is not given. We hypothesize that the extra annotation effort of counting the multiplicity of the concept of interest may help in front of more abstract or complex concepts. 


In this article we tackle the problem of indirect learning as "learning to count" by means of a deep neural network. Deep neural networks are particularly well suited for feature discovery. The proposed architecture relies on two different stages. At the first stage a convolutional network is used. We expect this stage to capture discriminative information about the concept of interest. The output of this stage is plugged in to a fully connected stage, where we expect the network will learn to count the multiplicity of the concept of interest. Figure \ref{fig:learning_to_count_concept} illustrates the proposal at a glance in the toy problem of representing even handwritten digits. The input image contains a random set of handwritten digits. In this problem, our goal is to represent even numbers by giving the amount of even numbers present in the image to the system. We explore the quality of the learned representation in two different tasks, classification of even and odd numbers, and digits recognition. We further visualize and analyze the learned concepts by means of a hierarchical approach. 

Our contributions are as follows: We introduce the problem of object representation as an indirect learning problem casted as a learning to count approach. Then, we validate the proposal in two scenarios. First, we learn to count even handwritten digits using the MNIST data set. And second we count pedestrian on a synthetically generated set of images from the UCSD pedestrian dataset. We then try to understand the representations that are learned for this task: Are CNN using object detectors or are they finding surrogate features? If this is the case, can "learning to count" be seen as an indirect, cheaper way of training object detectors? 

\cvprsection{Background and related works}
Our proposal is related to different techniques. On one hand we may find work on multiple instance learning related to similar concepts. Multiple instance learning \cite{Foulds2010} is a variation of supervised learning where instead of having individually labeled examples, instances comes in bags. These bags contain multiple instances. A bag is labeled positive if there is at least one example with the concept of interest, or labeled negative otherwise. The positive bag can be regarded as a set of attracting instances and the negative one as a set of repulsive instances. The classifier tries to learn a common concept across all the instances of the positive bag. As we commented in the introduction this approach is commonly found in large scale computer vision under the name of weakly supervised learning. 

Weakly supervised learning has been since widely used in computer vision since its inception and popularization \cite{Weber2000}, \cite{Fergus2003}. Early works used this in an standard instantiation of the MIL framework for inferring difficult to describe classes such as in \cite{Todorovic2006} where photometric, geometric, and topological features are recognized. More recently, several works, such as\cite{Nguyen2009}, explore the capacity of this technique for simultaneous localization and recognition. 

In multiple instance learning we find another approach closely related to our work, count-based multiple instance learning\cite{Foulds2010}. In count-based MIL the positive bag is composed of instances where the concept appears within the range of an interval. For example, the positive bag may contain images with 5 to 10 appearances of pedestrians. The main difference between this scenario and our proposal lies in the fact that even in count-based multiple instance learning the problem is casted as a binary problem. Either one belongs to the positive bag or to the negative one. In our case this distinction does not apply and the precise number of instances is the important value. This means that we do not explicitly define any set of negative samples. 

Because we are casting the problem of object description as learning to count, it is worthy to check the similarities with counting strategies in literature. From this point of view, counting has been tackled in different domains, such as cell counting \cite{Flaccavento2011} or pedestrian counting \cite{Lempitsky2010} \cite{Chan2008b}. The most common approach for counting firstly segments the objects of interest and then appropriate features are manually defined, automatically extracted, and then a supervised learner infers the final counting value \cite{Flaccavento2011}, \cite{Chan2008b}. Contrary to other approaches where bounding boxes or prototypes of the object of interest are used, in \cite{Lempitsky2010} the annotation tasks is reduced to dotting (pointing) and the counting process is casted as a problem of estimating image densities.
From the point of view of counting approaches and to the best of our knowledge the proposed approach is the first one where the counting problem is handled by learning deep features. Additionally, different from previous approaches we are not giving any hint on the object we are counting besides the occurrence multiplicity.

\cvprsection{Learning to count with CNN}
The main hypothesis of this work is that the multiplicity of the instances of the concept of interest in an image provides strong information regarding their discriminability for a feature learning process to exploit. CNN provide a nice framework for this problem since they naturally handle feature learning \cite{LeCun1989} and have shown impressive classification performance on different benchmark problems \cite{Krizhevsky2012}. 

Learning the multiplicity of the concept of interest is a counting problem. In this scenario, we want the architecture of the classifier to decouple the counting effect from the concept representation. For this purpose we may define a two block architecture with explicit semantics. The bottom stage is modelled using an agnostic architecture with a set of convolutional layers. Agnostic in the sense that there are little prior knowledge involved with respect to the concepts of interest. We expect this stage to capture and represent the underlying concept we are counting. The top stage comes in the form of fully connected layers with the goal of counting the occurrences of the concept. Thus, the last layer of the CNN has to output a number. In general we can cast this problem as either as a multi-class problem, a multi-label problem, or a regression problem. In those scenarios where the multiplicity of the concept of interest is small we may use multi-class or multi-label approaches, while when the number of occurrences is large regression strategies can be better suited. 

We consider networks of two or more convolutional layers followed by one or more fully connected layers. Each convolutional layer consist of several elements: a set of convolutional filters, ReLU non-linearities, max pooling layers and normalization layers. 

\begin{figure}[!ht]
\centering
\begin{tabular}{cc}
\includegraphics[width=4.0cm,height=4.0cm]{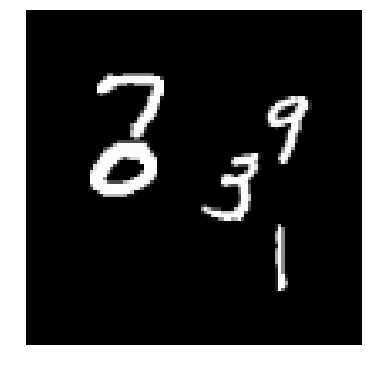}&\includegraphics[width=4.0cm,height=4.0cm]{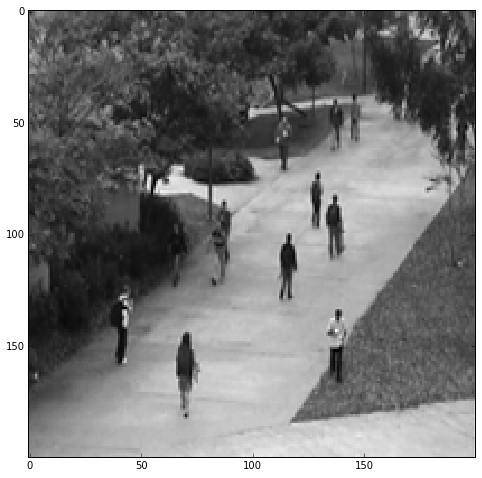}\\
(a)&(b)\\
\end{tabular}
\caption{Images from the two problems. (a) Training image for the learning to count even handwritten digits task, (b) Synthetic image generated from the UCSD data set for the task of counting pedestrians.}
\label{fig:datasets}
\end{figure}

In order to illustrate the concepts detailed in this article two different experiments are designed. First, we generate a synthetic problem based on the well known MNIST handwritten database. This first problem will serve to illustrate the basic ideas behind the article. We consider the task of even digits counting. Images containing up to five digits in arbitrary locations are generated. The training process will have as target label the amount of even digits present in the image. In this experiment we try to understand how the learner counts and what data representations are learnt. In the second experiment we use the proposed architecture for counting people from a surveillance camera recording. Because of scarce image availability the learner is trained on synthetic images. In this example we explore how learning to count allows for implicit target location. Additionally, we introduce a visualization tool for checking the image parts used for these tasks. Figure \ref{fig:datasets} shows an example of images for each problem.

\cvprsection{Experiments and results}

\cvprsubsection{Learning to count in the digits domain}
We focus on the problem of learning to count even handwritten digits. For this task data is labeled with the amount of even digits present in the image. We synthetically create a set of one million images of size $100\times 100$ including random digits from the MNIST database. A maximum of five digits are present in the image. The images are created with controlled overlapping by ensuring that two different numbers are 28 pixels away from each other, i.e. the distance between two digits centers is larger than 28 pixels. For this problem we use a four layers architecture CNN, shown in Table \ref{tab:network1}.

\begin{table}[!ht]
\centering
\begin{tabular}{c|c|c|c}
 Conv1 &  Conv2 & FC1 & FC2  \\ \hline
10x15x15 & 10x3x3 & 32 & 6 \\
x2 pool & x2 pool & & \\ \hline
\end{tabular}
\caption{CNN architecture for digits}
\label{tab:network1}
\end{table}

The algorithm is trained using the Caffe package\cite{Caffe} on a GPU NVIDIA Tesla K40. The network has been set to 300,000 iterations. 
The output layer is configured as a classification problem. 

\begin{figure}[!ht]
\begin{tabular}{cc}
\includegraphics[width=8.5cm]{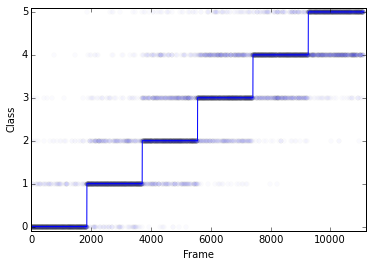}
\end{tabular}
\caption{Counting spread plot of the error in the even digits counting problem.}
\label{fig:mnist_svr}
\end{figure}

The accuracy of the base network is $93.8\%$. A spread plot of the error is shown in Figure \ref{fig:mnist_svr}. Observe that most of the frames are correctly labeled. Moreover, most of the errors correspond to adjacent number values.

Our hypothesis throughout this article considers the counting process as a surrogate task to potentially extract or infer interesting object descriptors. In order to assess the validity of our hypothesis we may check the performance of the learned representations across the different layers of the network on two different but related problems. First, we check the performance of the network on the even-odd classification task for individual digits. In this task the classifier is exposed to a single digit and its goal is to predict whether the digit is an even or odd number. Table \ref{tab:mnist_tab} shows the accuracy values if we train a support vector machine with the representations learned on different layers of the network. Observe that even though the original network task is to count the number of even numbers, the accuracy on the even-odd recognition considering the output of the convolutional layers and after the first fully connected layer are very high. This reinforce the idea that learning to count a certain concept implicitly extract discriminant features with respect to the concept we are counting. 

A second interesting question we could ask is how well these descriptors represent individual digits, i.e. does the network learn a representation for digit recognition? The second row of Table \ref{tab:mnist_tab} shows the test accuracy after training a multi-class support vector machine\footnote{Support vector machine is trained using three-fold cross-validation with rbf kernel. Parameters $C$ and $\gamma = 1/\sigma^2$ are set by means of a cross-validated grid search on a logarithmic scale. $C$ ranges from $10^{-1}$ to $10^{-3}$ and $\gamma$ from $10^{-5}$ to $10^{-8}$} using the same representations learned by the counting network. Even if the task is more different than the original question, the accuracy values obtained are very high. This result illustrates the potentiality of the technique and suggests that for the task of even digits counting, number identification plays an important role. Thus, the network is able to extract features that are not only important for the even number counting but also for even-odd recognition, and, furthermore, for individual number description. This result is similar to results observed in \cite{Zhou2015} where the authors found that in the task of scene recognition meaningful object detectors were also learned.

\begin{table}[!ht]
\centering
\begin{tabular}{|c|c |c| c|}
  \hline
 & Pool 1 & Pool 2 & FC 1 \\
 \hline
 Even vs. Odd & 0.929 & 0.983 & 0.992 \\
  \hline
 Numbers & 0.850 & 0.943 & 0.972 \\
  \hline
\end{tabular}
\caption{Accuracy performance in the even-odd recognition and in individual digits recognition problems using the descriptors learned for counting even numbers. The results are reported when using the representations after the first and second pooling layers of the CNN as well as after the first fully connected layer. }
\label{tab:mnist_tab}
\end{table}

It is worth analyzing the confusion matrix in the number recognition problem. $82.4\%$ of confusions correspond to numbers belonging to the same "superclass", either even or odd classes. On the contrary, only $17.6\%$ corresponds to confusions between even and odd numbers. This result shows that there is a bias on the individual digit identification, most probably corresponding to that of the original "learning to count" task. It is reasonable to think that the network focuses most of its effort in distinguishing between even and odd numbers and it does not really need to invest effort discriminating between elements of the even class.

\cvprsubsection{Learning to count pedestrians with deep features}
For the second experiment we use the UCSD pedestrian database \cite{Chan2008b}. Due to the small amount of annotated data we create set of 200.000 synthetic images containing different number of pedestrians. The creation of the images is as follows: First, a simple background subtraction technique is used. We use the labeled and unlabeled images of the database. For each video the pixel wise median is used as background. Then a Gaussian filter is used across the backgrounds to smooth out motion effects. From each video pedestrian candidates are extracted subtracting the background. Background noise and imperfect shapes are handled using mathematical morphology operators. A set of two hundred different pedestrian subimages are extracted. New images are synthesized by means of a composition with feathering using the region of interest from \cite{Chan2008b} that delineates the pedestrian track. Each image contains up to twenty-five pedestrians.

In this problem we are interested in counting the amount of pedestrians. We use a five layers architecture CNN with two convolutional layers followed by three fully connected layers (see Table \ref{tab:network2}). The algorithm is trained using the Caffe package\cite{Caffe} on a GPU NVIDIA Tesla K40. The network has been set to 50,000 iterations. 

\begin{table}[!ht]
\centering
\begin{tabular}{c|c|c|c|c}
 Conv1 &  Conv2 & FC1 & FC2 & FC3 \\ \hline
8x9x9 & 8x5x5 & 128 & 128 & 25 \\
x2 pool & x1 pool & & &\\ \hline
\end{tabular}
\caption{CNN architecture for pedestrians}
\label{tab:network2}
\end{table}

\begin{figure}[!ht]
\begin{tabular}{cc}
\includegraphics[width=8.5cm]{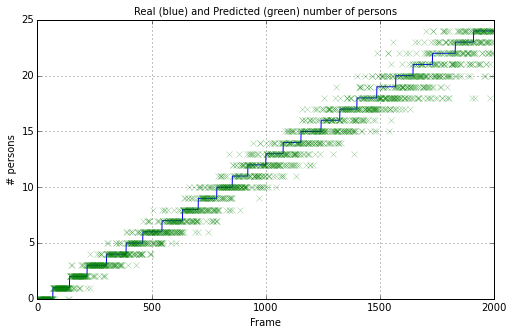}
\end{tabular}
\caption{Counting spread plot of the error in the pedestrian counting problem.}
\label{fig:ucsd_svr}
\end{figure}

The performance of the base network is $0.74$ mean absolute error and $1.12$ mean squared error. A spread plot of the error is shown in Figure \ref{fig:ucsd_svr}. The network closely follows the target values. Observe that the error deviation increases with the counting value, i.e. there is more confusion in more crowded scenes.

\cvprsubsection{Visualization and localization of concepts}
In order to visually check and verify what concepts are being learnt by the network we use a variation of the hypercolumns visualization \cite{Hariharan2014}. In the original visualization each pixel in the original image is described by all activations above it, undoing subsampling and pooling steps. These descriptions are fed to a grid of classifiers one per pixel (or group of pixels). The class with higher confidence at each pixel defines a map of the plausible locations that are active for a certain class. 

In the same spirit as the aforementioned work, we define a similar approach. Starting with the hypercolumn representation on the last layer we cluster the resulting hypercolumns into a set of prototypes using an online k-means algorithm. Then, a MIL approach with positive and negative instances with the concept of interest is used. In the case of even digits counting, the positive set consists of examples with even digits and the negative set contains examples with odd digits. Each image is then represented as a feature-normalized histogram of the of the occurrences of each word prototype for each image. Then a feature selection process is used. To this end, we propose to use a sparse selector such as an $\ell 1$-regularized linear support vector machine. Because all feature values are positive, positive values in the weights of the model correspond to discriminant features in favor of the positive class. By selecting the words with positive values and visualizing their location we may check the image support at that layer. In order to refine the visualization the process is repeated for the next layer. However, we condition the method on the positive areas of the previous stage. Therefore,a new clustering and prototype selection is performed. But this time, only the positive areas from the positive bag are jointly considered with the negative examples. The rest of the methodology remains the same.

\begin{figure}[!ht]
\centering
\begin{tabular}{c}
 \includegraphics[width=8cm]{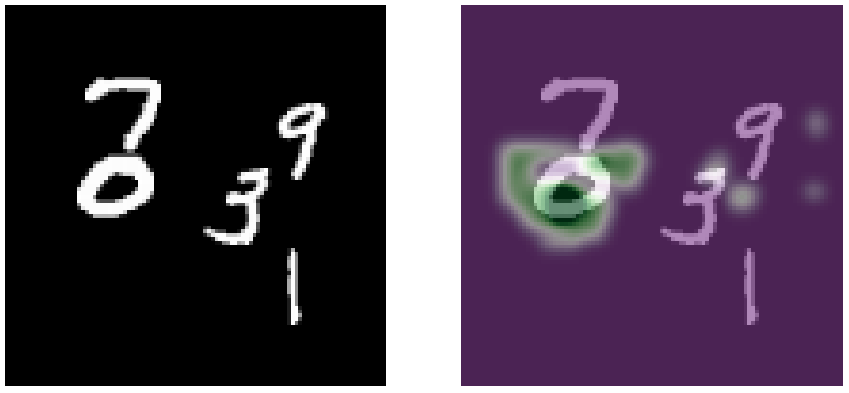}\\
 \includegraphics[width=8cm]{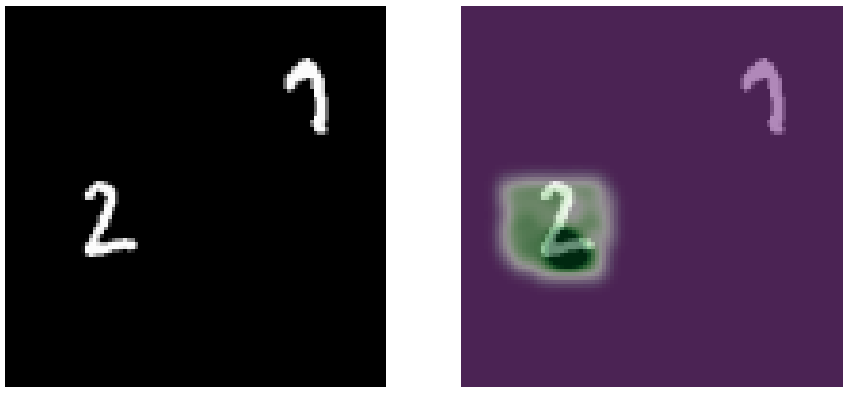}\\
 \includegraphics[width=8cm]{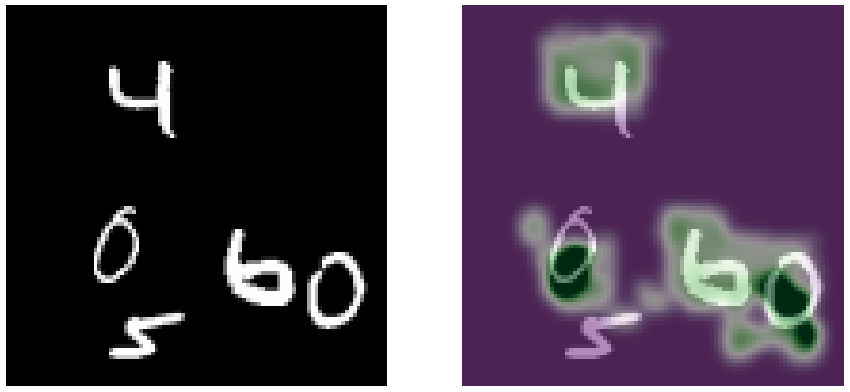}\\
 \includegraphics[width=8cm]{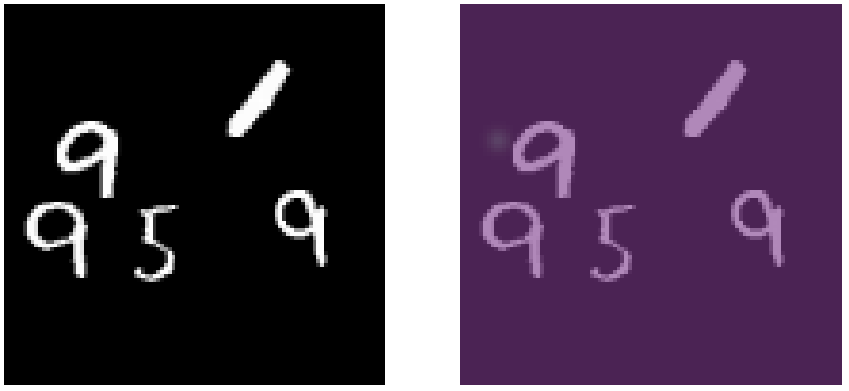}\\
\end{tabular}
\caption{Visualization of the location of the concept of interest. Green values correspond to higher activations. Observe that the network specifically fires on even digits.}
\label{fig:visualization1}
\end{figure}
\begin{figure}[!ht]
\centering
\begin{tabular}{c}
 \includegraphics[width=8cm]{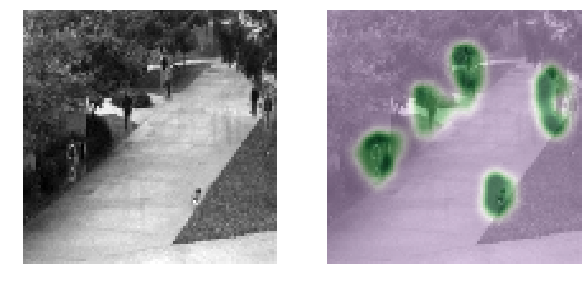}\\
 \includegraphics[width=8cm]{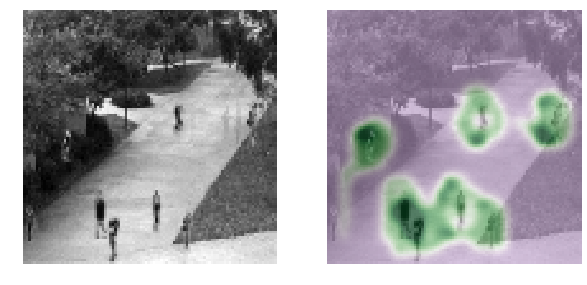}\\
 \includegraphics[width=8cm]{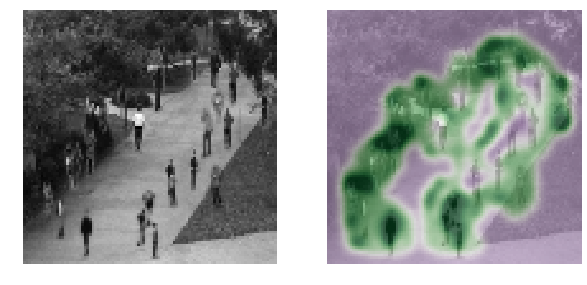}\\
\end{tabular}
\caption{Visualization of the location of the pedestrians in several images. Green values correspond to higher activations.}
\label{fig:visualization2}
\end{figure}

Figure \ref{fig:visualization1} and \ref{fig:visualization2} show some examples of the visualization and localization of the responses of the network according to the proposed methodology for the digits and pedestrians problems. We may see that the concepts of interest are correctly located. Moreover, the specific discriminant parts are colored in green.

\cvprsection{Conclusions}
In this paper we explore the task of counting occurrences of a concept of interest with CNN. Additionally, we introduce the problem of object representation as an indirect learning problem casted as a learning to count approach. The proposal is illustrated in two synthetic scenarios: learning to count even handwritten digits in an image and counting the number of pedestrians form a surveillance camera. We transfer the representation learned in the  problem of even digits counting to two different recognition problems, namely, even-odd classification of individual digits and digits recognition. The performance results on these problems is high. This suggests that the task of counting even digits may be used as a surrogate for finding good representations for these new tasks. We further analyze our proposal in a pedestrian counting scenario. The results in this scenario are encouraging and reinforce the feasibility of the proposal in front of counting problems.

In our opinion this article opens an interesting line of research that deserve further attention and study. There are still many open questions to be addressed such as, which is the best network architecture for counting? or when is a counting task a good surrogate for recognition problems?

\cvprsection*{Acknowledgments}
This work has been partially funded by the Spanish MINECO Grants TIN2013-43478-P and TIN2012-38187-C03. We gratefully acknowledge the support of NVIDIA Corporation with the donation of a Tesla K40 GPU used for this research.
\end{document}